\pdfoutput=1

\documentclass[11pt]{article}

\usepackage[final]{acl}

\usepackage{times}
\usepackage{latexsym}

\usepackage[T1]{fontenc}

\usepackage[utf8]{inputenc}

\usepackage{microtype}
\usepackage{inconsolata}
\usepackage{graphicx}
\usepackage{subfigure} 
\usepackage{amsmath}
\usepackage{stfloats}
\usepackage{multirow}
\usepackage{enumitem}
\usepackage{booktabs}
\usepackage{bbding}
\usepackage{amsfonts,amssymb}
\usepackage{diagbox}
\usepackage{newfloat}
\usepackage{listings}
\usepackage{times}  
\usepackage{helvet} 
\usepackage{courier}  
\usepackage{natbib}  
\usepackage{caption}
\usepackage{makecell}
\usepackage{xcolor}
\usepackage{pifont}
\usepackage{soul}
\usepackage{xcolor}
\usepackage{url}
\usepackage{tikz}
\usepackage{colortbl} 
\usepackage{CJKutf8} 
\usepackage[linesnumbered,ruled,vlined]{algorithm2e}

\definecolor{red}{RGB}{255,0,0}
\definecolor{green}{RGB}{0,255,0}

\title{\textsc{CRAT}: A Multi-Agent Framework for Causality-Enhanced Reflective and Retrieval-Augmented Translation with Large Language Models}

\author{
 Meiqi~Chen$^{1}$\footnotemark[1], 
 Fandong~Meng$^{2}$\footnotemark[2],
 Yingxue~Zhang$^{2}$,
 Yan~Zhang$^{1}$,  Jie~Zhou$^{2}$\\ 
 $^1$Peking University
 $^2$Pattern Recognition Center, WeChat AI, Tencent Inc, China\\
\texttt{meiqichen@stu.pku.edu.cn},  \texttt{zhyzhy001@pku.edu.cn},\\
\texttt{\{fandongmeng, yxuezhang, withtomzhou\}@tencent.com} \\
}

\begin{document}
\maketitle
\begin{abstract}
Large language models (LLMs) have shown great promise in machine translation, but they still struggle with contextually dependent terms, such as new or domain-specific words. This leads to inconsistencies and errors that are difficult to address.
Existing solutions often depend on manual identification of such terms, which is impractical given the complexity and evolving nature of language. 
While Retrieval-Augmented Generation (RAG) could provide some assistance, its application to translation is limited by issues such as hallucinations from information overload.
In this paper, we propose \textsc{CRAT}, a novel multi-agent translation framework that leverages RAG and causality-enhanced self-reflection to address these challenges. This framework consists of several specialized agents: the Unknown Terms Identification agent detects unknown terms within the context, the Knowledge Graph (KG) Constructor agent extracts relevant internal knowledge about these terms and retrieves bilingual information from external sources, the Causality-enhanced Judge agent validates the accuracy of the information, and the Translator agent incorporates the refined information into the final output. This automated process allows for more precise and consistent handling of key terms during translation.
Our results show that \textsc{CRAT} significantly improves translation accuracy, particularly in handling context-sensitive terms and emerging vocabulary. 
\end{abstract}

\renewcommand*{\thefootnote}{\fnsymbol{footnote}}
\footnotetext[1]{Work was done during an internship at Pattern Recognition Center, WeChat AI, Tencent Inc.}
\footnotetext[2]{Corresponding author.}

\renewcommand*{\thefootnote}{\arabic{footnote}}
\setcounter{footnote}{0} 

\section{Introduction}

Large language models (LLMs), such as GPT-4~\cite{openai2023gpt4} and Llama~\cite{Hugo2023LLaMA}, are increasingly becoming foundational for NLP tasks. Recently, several efforts have been made to leverage the capabilities of these models for machine translation~\cite{hendy2023good, jiao2023chatgpt, zhang2023prompting, wang2024deltaonlinedocumentleveltranslation}, achieving promising results that demonstrate their potential in practical applications.

\begin{figure}[t]
\centering  
\includegraphics[width=0.48\textwidth]{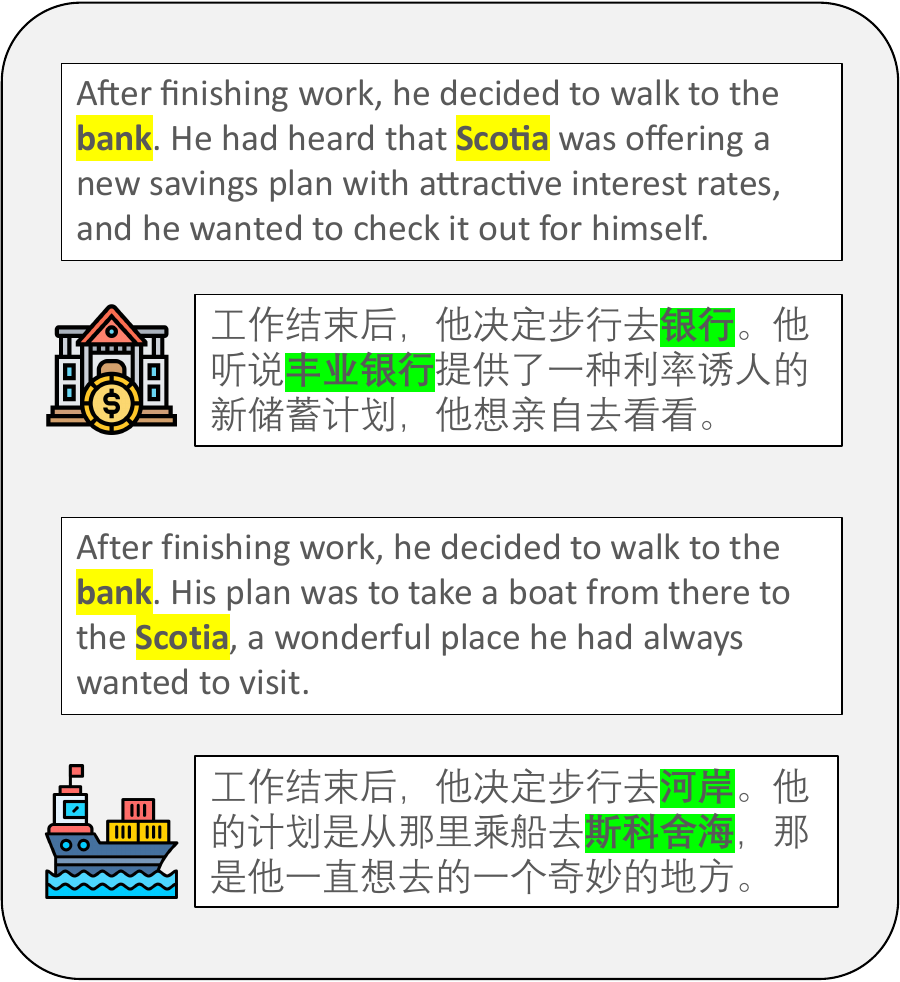}
\caption{
\begin{CJK*}{UTF8}{gbsn} 
Demonstration of the contextual ambiguity LLMs face during English-Chinese translation. In the top example, ``\emph{bank}'' and ``\emph{Scotia}'' are interpreted within a financial context, while in the bottom example, they shift to a geographical meaning. This illustrates the necessity for combining both internal knowledge and external, domain-specific information to accurately reason about context-dependent terms.
\end{CJK*}
}
\label{fig:example}
\end{figure}

However, using LLMs for translation heavily relies on the model's generative abilities. This is because LLMs passively learn to model the relationships between contexts based on pretraining corpora~\cite{cao2022can, sun2024causal}. When a word no longer carries its usual meaning in a particular context, LLMs tend to struggle, and it becomes challenging to maintain consistency in translating entities and referential terms. As shown in Figure~\ref{fig:example}, where the terms ``\emph{bank}'' and ``\emph{Scotia}'' have different meanings depending on the context. In the first case, these terms refer to a financial institution and its services, while in the second, they shift to a geographical and navigational context. This contextual ambiguity demonstrates the need for LLMs to access both internal knowledge (from the original context) and external knowledge (updated or domain-specific information) to accurately infer and translate terms based on the situation.

This challenge underscores the importance of adapting LLMs to practical translation scenarios. The key to solving this problem lies in recognizing special terms that need attention and determining their precise meanings. Previous methods typically rely on researchers' prior knowledge to manually identify these terms, including acronyms, new words, slang, or borrowed words~\cite{fiederer2009quality, CHOI2017149, vaibhav2019improving}. However, given the complexity and diversity of real-world data, manually identifying terms one by one is impractical. Furthermore, the dynamic nature of language leads to the constant emergence of new word meanings, making it challenging for LLMs to translate accurately, especially when the models have not been updated.

Thus, there is a pressing need for an automated approach to identify and clarify the meanings of such terms. 
While retrieval-augmented generation (RAG) has been applied to question-answering (QA) tasks~\cite{ye2021rng, asai2023self}, it poses challenges when used directly for translation. Excessive external information can interfere with accurate translation, leading to errors or hallucinations~\cite{gao2023enabling, ding2024retrieve}.

To address this issue, in this paper, we propose \textsc{CRAT}, a multi-agent framework that leverages the powerful instruction-following and tool-using capabilities of LLMs for \textbf{C}ausality-Enhanced Reflective and \textbf{R}etrieval-\textbf{A}ugmented \textbf{T}ranslation.
This framework allows LLMs to autonomously identify unknown terms and clarify their meanings within the given context. Specifically, an \textbf{Unknown Terms Detector agent} identifies unknown terms in the context. Next, a \textbf{Knowledge Graph (KG) Constructor agent} extracts internal knowledge about these terms from the context and retrieves relevant information from external sources, to construct a KG for translation (a.k.a., TransKG). Then, a \textbf{Causality-enhanced Judge agent} uses causality-enhanced reflection to evaluate whether the retrieved information is relevant and appropriate, facilitating the construction of TransKG. Finally, a \textbf{Translation agent} references the updated information to produce a precise and consistent translation.

Overall, our main contributions are as follows: 
\begin{itemize}
\item  We introduce a novel framework for retrieval-augmented machine translation, which could automatically gather internal and external information on unknown terms to construct a TransKG, reducing the reliance on manual intervention.
\item We make the first attempt to apply a causality-enhanced reflection mechanism for retrieval-augmented machine translation, which ensures contextual meanings and improves translation accuracy.
\item Experimental results demonstrate that our framework significantly improves translation accuracy and consistency.
\end{itemize}
\section{Related Work}
\subsection{LLMs for Machine Translation}
Recent work on LLMs for machine translation highlights both their potential and their limitations, particularly when dealing with context-dependent or domain-specific terms~\cite{peng2023towards, hendy2023good, jiao2023chatgpt, zhang2023prompting}. For example, \citet{he2024exploring} explores human-like translation strategies by encouraging LLMs to imitate human reasoning, which improves their ability to handle nuanced contexts. Similarly, \citet{Zeng_Meng_Yin_Zhou_2024} and \citet{wang2024taste} introduce self-reflection mechanisms that enable LLMs to assess and refine their own translations, enhancing overall accuracy. However, these approaches still rely heavily on the model’s pretraining and may struggle with newly emerging terms or ambiguous contexts.
\citet{ghazvininejad2023dictionary} investigate dictionary-based prompting to improve phrase-level translation, demonstrating the value of incorporating external knowledge into the translation process. While effective, this method requires structured input, which can limit its flexibility in more dynamic scenarios. Our work overcomes this by integrating both internal and external knowledge sources dynamically through a multi-agent framework, automating the identification and clarification of unknown terms.

\subsection{Retrieval Augmented Translation}
RAG has been widely applied in tasks such as question answering~\cite{ye2021rng, asai2023self}, but its use in translation remains relatively limited. One major challenge is hallucination, where excessive retrieval of information introduces inaccuracies. \citet{zhang2024enhancing} applied RAG to the translation of e-commerce product titles, successfully improving accuracy by retrieving domain-specific information. \citet{conia2024towards} introduce a RAG framework for multilingual machine translation, specifically targeting domain-specific translations. 
However, these approaches still risk overloading the translation with irrelevant data. Similar challenges are noted in \cite{gao2023enabling, yan2024corrective, ding2024retrieve, chen2024improving}, where feeding retrieval information directly to LLMs without verification often results in hallucinations.

\subsection{Causality-aware Generation}
Some studies have pointed out that LLMs may produce inaccurate output due to reliance on spurious correlations learned from pretraining corpus~\cite{cao2022can, sun2024causal}. 
As for the translation task, \citet{chesterman2017causal} propose a causal model where human translations are explicitly seen both as caused by antecedent conditions and as causing effects on readers and cultures. \citet{ni2022original} studies how ``translationese'' (text translated by humans) affects machine translation performance through causal analysis.
However, to the best of our knowledge, we are the first work to apply causal reasoning mechanisms to LLMs for machine translation tasks. By leveraging causal invariance\cite{pearl2009causality, pearl2018book}, our approach validates the retrieved information and promotes more accurate translations, particularly for terms and concepts that are ambiguous or not often encountered during pretraining.
\begin{algorithm*}[t]
    \SetAlgoLined
    \SetAlgoNoLine
    \caption{CRAT Inference}\label{algo:framework}
    \SetKwInOut{Require}{Require}
    \SetKwInOut{Input}{Input}
    \SetKwInOut{Output}{Output}
    \Require{$U$ (Unkown Terms Identification), $K$ (Knowledge Graph Constructor), $C$ (Causality-enhanced Judge), $T$ (Translator)}
    \Input{$x$ (Input context)}
    \Output{$y$ (Generated translation)}
    $U$ identifies $M=\left\{m_1, \ldots, m_{N1}\right\}$\\
    $K$ extracts Internal\_Knolwedge $k_{I}$ related to $M$ from $x$\\
    $K$ retrieves documents $D = \{d_1, d_2, ..., d_{N2}\}$ from external sources\\
    \textbf{Confidence} = Give a final judgment based on the relevance of each pair ($k_{I}$, $d_i$), $d_i \in D$ \\
    \tcp{\textbf{Confidence} has 2 optional values: [CORRECT] or [INCORRECT]} \label{cmt} \leavevmode
    \uIf{$\mathrm{Confidence}_{i}$ == \texttt{[CORRECT]}}{
    External\_Knowledge $k_{E}$ += $d_{i}$
    }
    $T$ predicts $y$ given $x$, $k_{I}$, and $k_{E}$ 
\end{algorithm*}

\section{Method}
\begin{figure*}
\centering  
\includegraphics[width=1.0\textwidth]{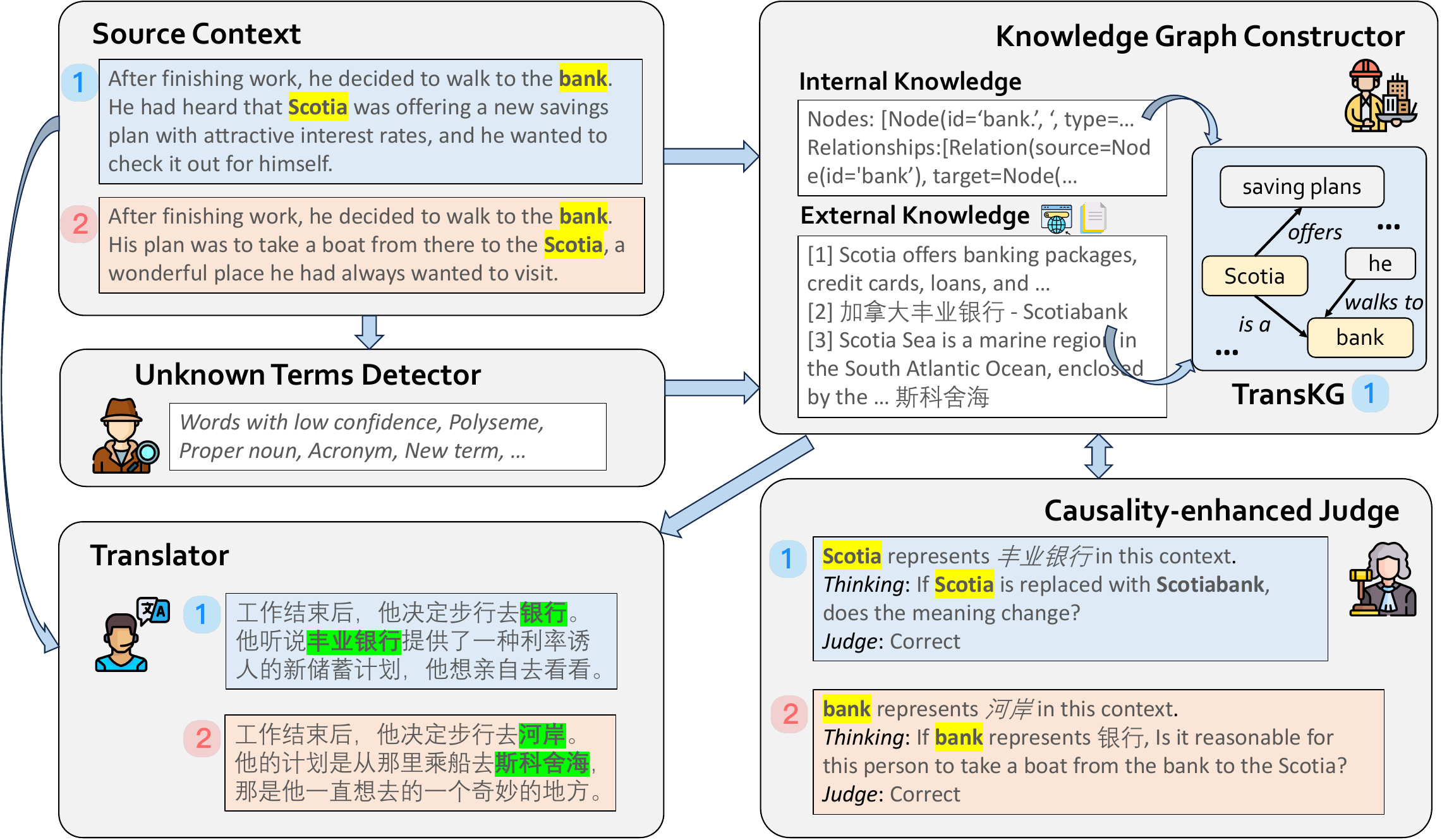}
\caption{Our proposed multi-agent translation framework (i.e., CRAT) for improving LLM translation accuracy, especially for unknown terms. The Unknown Terms Detector Agent identifies terms with low confidence or ambiguity, such as polysemes or new terms. The Knowledge Graph Constructor Agent then builds a TransKG based on both internal and external sources to collect the necessary information. The Causality-enhanced Judge Agent evaluates the appropriateness of term meanings by considering causal invariance in context. Finally, the Translator Agent uses accurate meanings to generate translations.}
\label{fig:method}
\end{figure*}

In this section, we outline the proposed multi-agent translation framework (i.e., CRAT) that leverages the Retrieval-Augmented Generation (RAG) mechanism and causality-enhanced self-reflection. The framework addresses the challenge of translating uncertain terms with accuracy and contextual relevance, overcoming limitations observed in LLMs when encountering such terms. The overall pipeline is presented in Figure~\ref{fig:method}.
\subsection{Task Formulation}
As shown in Algorithm~\ref{algo:framework}, given input source context $\mathcal{X}$, the framework is expected to generate the translation $\mathcal{Y}$.
The entire framework typically includes an unknown terms detector $\mathcal{U}$, a knowledge graph constructor $\mathcal{K}$, a causality-enhanced judge $\mathcal{C}$, and a translator $\mathcal{T}$. The detector $\mathcal{U}$ identifies unknown terms $\mathcal{M}=\left\{m_1, \ldots, m_{N1}\right\}$ from the input $\mathcal{X}$. The constructor $\mathcal{K}$  extracts internal knowledge $\mathcal{K}_{I}$ related to the unknown terms $\mathcal{M}$ from the input $\mathcal{X}$ and retrieves relevant documents $\mathcal{D} = \left\{d_1, \ldots, d_{N2}\right\}$ from external sources. The judge $\mathcal{C}$ then evaluates the relevance of each document and determines the refined external knowledge $\mathcal{K}_{E}$. Based on the input $\mathcal{X}$ and the knowledge $\mathcal{K}_{I} + \mathcal{K}_{E}$, the translator $\mathcal{T}$ finally generate the translation $\mathcal{Y}$. This framework can be formulated as:
\begin{equation}
\small
P(\mathcal{Y} \mid \mathcal{X})= P(\mathcal{K}_{I} \mid \mathcal{X})P( \mathcal{K}_{E} \mid \mathcal{X}) P(\mathcal{Y} \mid \mathcal{X}, \mathcal{K}_{I}, \mathcal{K}_{E})
\end{equation}

\subsection{Unknown Terms Detector}
The Unknown Terms Detector agent is responsible for identifying terms within the source context that may pose challenges for accurate translation. These terms typically include words or phrases with low confidence by agent, polysemes (words with multiple meanings, e.g., ``\emph{bank}'' in Figure~\ref{fig:method}), acronyms (e.g., F.B.I for Federal Bureau of Investigation), proper nouns (e.g., ``\emph{Sccotia}'' in Figure~\ref{fig:method}), or new terms that the agent may not fully comprehend. By autonomously identifying these terms, the agent highlights them for further analysis, ensuring that they receive special attention during the translation process. This step is crucial for maintaining translation accuracy, as it allows the framework to proactively address potential ambiguities or inconsistencies before the translation is finalized.

\subsection{Knowledge Graph Constructor}
The Knowledge Graph Constructor plays a crucial role in enhancing translation accuracy by collecting internal and external knowledge related to the detected unknown terms, thereby resolving contextual ambiguities.
Specifically, it constructs a Translation Knowledge Graph (TransKG), a structured representation of term relationships and meanings based on internal and external knowledge.

\paragraph{Internal Knowledge Extraction} The agent begins by identifying and structuring internal knowledge directly from the source context. It identifies relationships between terms, such as (``\emph{Scotia}'', ``\emph{offers}'', ``\emph{savings plan}'') in the first example shown in Figure~\ref{fig:method}.
By organizing these relationships into nodes and edges within a graph, TransKG provides a structured representation that captures the context-specific details necessary for translation.

\paragraph{External Documents Retrieval} 
We also enable the agent to access external sources, such as online databases or other external corpora. It retrieves domain-specific and bilingual information, such as services provided by ``\emph{Scotiabank}'' or geographical details of the ``\emph{Scotia Sea}'', which are crucial for disambiguating terms with multiple meanings or unfamiliar references.

\paragraph{Contextual Knowledge Integration} 
After validation, the agent could integrate that relevant and appropriate information into TransKG by linking it to the relevant nodes, thereby enriching the contextual understanding --- e.g., associating (``\emph{Scotia}'', ``\emph{is a}'', ``\emph{bank}'') as shown in Figure~\ref{fig:method}. 

This unified graph offers a comprehensive view of the relationships and meanings of terms, enabling us to accurately interpret and resolve ambiguities. TransKG’s dynamic capabilities ensure that the knowledge graph remains relevant and updated, incorporating the latest information needed for accurate translation. The validation process will be detailed in Section~\ref{subsec:causal}.

\subsection{Causality-enhanced Judge}
\label{subsec:causal}

Following the construction of TransKG, the Causality-Enhanced Judge agent evaluates whether retrieved information is contextually relevant and whether ambiguous terms are accurately interpreted based on their intended meanings. For example in Figure \ref{fig:method}, terms like ``\emph{bank}'' and ``\emph{Scotia}'' can carry multiple possible meanings depending on context. To ensure causal invariance~\cite{pearl2009causality, pearl2018book}, the agent uses a causality-driven reflection mechanism, testing whether substituting these terms in translation preserves their original semantic integrity.

The process begins after the framework identifies unknown or ambiguous terms and gathers internal and external knowledge about them. The Causality-enhanced Judge evaluates the contextual implications of these terms, examining if alternative interpretations or translations align with the overall narrative or cause a shift in meaning. 
Causal invariance demands that, if terms like ``\emph{Scotia}'' are translated to imply either a financial institution or a geographical location, back-translating  into the original language should yield the same conceptual alignment. This can be further clarified through counterfactual reasoning~\cite{pearl2009causality,peters2017elements}:
if the back-translation introduced unintended meanings, this would lead to a misalignment in understanding. In such cases, the judge discards the retrieved information (i.e., assess them as \texttt{[INCORRECT}]), ensuring that only semantically stable knowledge is added to TransKG.

In this way, the Causality-enhanced Judge agent ensures that the external knowledge retrieved is relevant and accurate, avoiding models' reliance on incorrect or outdated information. This agent also enhances the ability to maintain consistency and accuracy in translation, especially when dealing with polysemy, proper nouns, or emerging terms, etc. By leveraging causality, our framework could verify that the target meaning of unknown terms is logically sound within its context, minimizing the risk of errors and misunderstandings.

\subsection{Retrieval-Augmented Translator}

The Translator agent is the final component of the framework, responsible for producing the translated output using the refined and contextually accurate meanings provided by the earlier agents. Once the Causality-enhanced Judge validates and determines the interpretations of ambiguous terms, the Translator references this adjusted information to generate a precise and coherent translation.

The Translator does not rely solely on direct word-for-word translation. Instead, it incorporates contextual knowledge and causality-aware reflection from other agents to ensure the translation preserves both the semantic integrity and the original intent. For example, if the term ``\emph{bank}'' is identified as a geographical feature instead of a financial institution, the Translator will use the appropriate term in the target language (e.g., \begin{CJK}{UTF8}{gbsn} ``河岸'' rather than ``银行''\end{CJK} in Chinese) to convey the correct meaning in the given context.

This ensures that the translated output is not only linguistically accurate but also contextually relevant and logically coherent, providing a natural and fluent translation. By combining the verified meanings from the earlier steps, the Translator achieves a high level of consistency, reducing errors commonly associated with polysemous or context-dependent terms.
\begin{table*}
    \renewcommand
    \arraystretch{1.0}
    \centering
    \setlength{\tabcolsep}{10pt}
    \begin{tabular}{l|ccc|ccc}
    \toprule
         \multirow{2}{*}{\bf{Model}} & \multicolumn{3}{c|}{\textbf{$\mathrm{En} \Rightarrow \mathrm{Zh}$}} & \multicolumn{3}{c}{\textbf{$\mathrm{Zh} \Rightarrow \mathrm{En}$} }\\ 
         \cmidrule(lr){2-4}\cmidrule(lr){5-7}
         & BLEU   & COMET   & CONSIS & BLEU   & COMET   & CONSIS \\ \midrule
          GPT-3.5-turbo &27.6 &82.0 &82.1 & 41.9 &82.2 &81.3\\ 
          + \texttt{CRAT}&28.4 &82.9 &86.7  &43.5 &82.9 &84.5 \\ \midrule
          GPT-4o &32.7 &83.6 &83.5 &43.8 &82.4 &82.6 \\ 
          + \texttt{CRAT} &33.5 &84.5 &86.7 &46.0 &83.6 & 85.2\\ \midrule
          Qwen-7B-Instruct &27.6 &82.5 &81.2 &39.7 &79.8 &80.1\\ 
          + \texttt{CRAT} &29.2 &83.3 &83.8 &41.8 &81.0 &83.0\\ \midrule
          Qwen-72B-Instruct &29.9 &83.0 &81.7 &47.1 &82.4 &80.6\\ 
          + \texttt{CRAT} &30.8 &84.6 &85.0 &50.0 &83.2 &83.8 \\ 
    \bottomrule
    \end{tabular}
        \caption{Translation performance (\%) comparison across various LLMs on New York Times reports. The results are reported with and without the proposed CRAT framework for English-to-Chinese ($\mathrm{En} \Rightarrow \mathrm{Zh}$) and Chinese-to-English ($\mathrm{En} \Rightarrow \mathrm{Zh}$) tasks.}
    \label{tab:main1}
\end{table*}
\section{Experiments}
\subsection{Experimental Setup}
\paragraph{Dataset}
We collect news articles from the New York Times Chinese website \footnote{https://cn.nytimes.com/china/}, including both Chinese and English versions. We filter the data to only include reports from 2024 to minimize the likelihood that these reports were encountered during the training phase of LLMs. For each language pair, we retain 500 data points.

\paragraph{Baselines}
We implement each agent using an LLM and evaluate four LLMs in a zero-shot fashion. For limited-access LLMs, we choose two GPT-series models, \texttt{ GPT-3.5-Turbo-0125} and \texttt{GPT-4o}. We get access to these models through the official API provided by OpenAI and retain the hyper-parameters as default. For open-source LLMs, we deploy \texttt{Qwen-7B} and  \texttt{Qwen-72B}. The \texttt{max\_new\_tokens} is set to 2048 and other hyper-parameters remain default.

\paragraph{Metrics}
Following previous work, we adopt BLEU~\cite{papineni-etal-2002-bleu}, reference-free COMET~\cite{rei-etal-2022-cometkiwi}\footnote{ \texttt{Unbabel/wmt22-comet-da},
\url{https://github.com/Unbabel/COMET/}}, and an evaluator implemented using \texttt{GPT-4o}, which we refer to as CONSIS, to measure translation quality. BLEU and COMET provide quantitative assessments of translation performance --- BLEU by comparing n-grams between the translated and reference texts, and COMET by leveraging neural networks to predict human judgment. In contrast, the \texttt{GPT-4o}-based evaluator (i.e., CONSIS) is designed to assess the translation more holistically and offers a deeper semantic analysis. It focuses on whether the translation accurately conveys the original meaning, with particular focus on the accuracy and consistency of the unknown terms. 

\subsection{Main Results}
From Table~\ref{tab:main1}, we observe that:

(1) Across all models, the integration of CRAT consistently increases BLEU scores, demonstrating that CRAT positively impacts the models' alignment with reference translations.

(2) The enhancements in COMET and CONSIS scores indicate that  CRAT effectively enhances translation accuracy and consistency. The greatest performance enhancements are seen in \texttt{GPT-4o} and \texttt{Qwen-72B-Instruct}, suggesting that more advanced LLMs may better leverage CRAT's framework.

(3) Among the evaluated models, \texttt{GPT-4o} and \texttt{Qwen-72B-Instruct} with CRAT exhibit the highest scores across all metrics, demonstrating that the combination of an advanced model and the CRAT framework yields superior translation outcomes. In contrast, smaller models like \texttt{GPT-3.5-turbo} and \texttt{Qwen-7B-Instruct} show less pronounced improvements, which may be due to their more limited baseline capabilities.

\begin{table}
    \renewcommand
    \arraystretch{1.0}
    \centering
    \small
    \setlength{\tabcolsep}{2pt}
    \begin{tabular}{l|l|c|c|c}
    \toprule
     \bf{Id} & \bf{Setting} &\bf{BLEU} &\bf{COMET} & \bf{CONSIS}\\ 
         \midrule
       1 &Vanilla &29.9 &83.0 &81.7  \\
         \midrule
       2 &\makecell[l]{1 + (unrefined) TransKG)}  &30.4 &83.9 &83.6 \\
       \midrule
       3 &\makecell[l]{2 + Causality-enhanced \\ Judge (\texttt{CRAT})} &30.8 &84.6 &85.0 \\
    \bottomrule
    \end{tabular}
        \caption{Ablation study with \texttt{Qwen-72B-Instruct} on New York Times reports (\%) across three configurations: (1) Vanilla, (2)  Vanilla with (unrefined) TransKG, and (3) Our proposed CRAT. }
    \label{tab:ab}
\end{table}

\subsection{Ablation Study}
From Table~\ref{tab:ab}, configuration 1 denotes vanilla \texttt{Qwen-72B-Instruct}, configuration 2 denotes employing the Unknown Terms Detector and the Knowledge Graph Constructor to obtain an unrefined TransKG, and configuration 3 denotes the entire CRAT framework.
We observe that:

(1) Adding TransKG (Configuration 2) leads to improvements in all metrics compared to the Vanilla setup, showing that incorporating internal and external knowledge boosts translation quality and enhances consistency.

(2) Furthermore, Configuration 3 (i.e., our proposed CRAT), which 
 integrates the Causality-Enhanced Judge agent, showing the highest performance across all metrics. This demonstrates that the causality-enhanced reflection mechanism significantly improves not only the accuracy but also the contextual understanding and consistency of translations, underlining the effectiveness of the proposed CRAT framework.

\begin{table*}
    \renewcommand
    \arraystretch{1.0}
    \centering
    \setlength{\tabcolsep}{6pt}
    \begin{tabular}{l|c}
    \toprule
        \textbf{Model} & \textbf{Tranlation}  \\
         \midrule
        \textbf{Source Context} &\emph{\makecell[c]{
        Even sweeter concoctions await inside Lacaph,  a classy new coffeehouse \\ in District 1, just off \textbf{Rach Ben Nghe}, the slim urban canal that snakes \\through the city. Decorated with dark wood paneling and track lighting,\\ the cafe serves lemonade (80,000 dong) combined with coffee-blossom honey \\ and a dose of coffee brewed in a traditional Vietnamese \textbf{phin}.}}
  \\
         \midrule 

          GPT-4o & \begin{CJK}{UTF8}{gbsn}
\parbox[t]{0.5\textwidth}{
更甜美的饮品在 Lacaph 等着您，这是一家位于第一郡的新时尚咖啡馆，毗邻蜿蜒穿城而过的狭长城市运河——\textcolor{red}{{Rach Ben Nghe}}。咖啡馆装饰着深色木板墙和轨道灯，提供柠檬水（80,000越南盾），与咖啡花蜜和用传统越南\textcolor{red}{{滤杯（phin）}}冲泡的咖啡组合而成。
}
\end{CJK} \\ \midrule
          GPT-4o + \texttt{CRAT} & \begin{CJK}{UTF8}{gbsn}
\parbox[t]{0.5\textwidth}{
更甜的混合饮品在等着你, 在第一区新开的优雅咖啡馆Lacaph里。这家咖啡馆就在蜿蜒穿过城市的狭长运河\textcolor{blue}{\textbf{滨义河}}边。咖啡馆装饰着深色木板墙和轨道灯，供应柠檬水（80,000越南盾），混合了咖啡花蜂蜜和用传统越南\textcolor{blue}{\textbf{滴漏咖啡壶}}冲泡的咖啡。
}
\end{CJK} \\  \midrule \midrule
        \textbf{Source Context} &\emph{\makecell[c]{Every year, Apple and Google announce major software updates that \\ bring new features to our smartphones, like \textbf{cosmetic overhauls} to the \\ home screen, stronger privacy protections and fun messaging tools. \\This year, the changes will feel more radical because the companies \\ are focusing on reinventing our phones with artificial intelligence.
}}
  \\
         \midrule 
          Qwen-72B-Instruct & \begin{CJK}{UTF8}{gbsn}
\parbox[t]{0.5\textwidth}{
每年，苹果和谷歌都会宣布重大软件更新，为我们的智能手机带来新功能，如对主屏幕的\textcolor{red}{\textbf{美容翻新}}、更强的隐私保护以及有趣的即时通讯工具。而今年的变化会更加激进，因为这两家公司正致力于利用人工智能重塑我们的手机。
}
\end{CJK} \\ \midrule
          \makecell[c]{Qwen-72B-Instruct \\+ \texttt{CRAT}} &  \begin{CJK}{UTF8}{gbsn}
\parbox[t]{0.5\textwidth}{
每年，苹果和谷歌都会宣布重大软件更新，为我们的智能手机带来新功能，比如主屏幕的\textcolor{blue}{\textbf{外观调整}}、更强的隐私保护和有趣的即时通讯工具。今年，这些变化会让人感觉更加激进，因为这些公司正致力于用人工智能重塑我们的手机。
}
\end{CJK} \\ \midrule \midrule
\textbf{Source Context} &\emph{\makecell[c]{
Typhoon \textbf{Gaemi} (2024) (T2403, 05W, Carina) – a powerful \\typhoon that 
impacted East China.\\ \textbf{Gaemi} also drenched \textbf{western Luzon} in the Philippines.
}}
  \\
\midrule 
        Qwen-72B-Instruct & \begin{CJK}{UTF8}{gbsn}
\parbox[t]{0.5\textwidth}{
台风\textcolor{red}{\textbf{卡米}} (2024)（T2403，05W， Carina）是一场强烈的台风，影响了中国东部。\textcolor{red}{\textbf{盖米}}也给菲律宾\textcolor{red}{\textbf{西吕宋}}带来了大量降雨。
}
\end{CJK} \\ \midrule
          \makecell[c]{Qwen-72B-Instruct \\+ \texttt{CRAT}}&  \begin{CJK}{UTF8}{gbsn}
\parbox[t]{0.5\textwidth}{
台风\textcolor{blue}{\textbf{格美}} (2024)（T2403，05W，卡琳娜）是一场强烈的台风，影响了中国东部。\textcolor{blue}{\textbf{格美}}也给菲律宾\textcolor{blue}{\textbf{吕宋岛西部}}带来了大量降雨。

}
\end{CJK} \\

    \bottomrule
    \end{tabular}
        \caption{Case Study of GPT-4o and Qwen-72B-Instruct when dealing with ambiguous terms such as ``\emph{bank}'' and ``\emph{Scotia}''. Our \texttt{CRAT} framework shows improved accuracy in distinguishing different contextual nuances by integrating knowledge sources and causality-aware reflections, ensuring more consistent and precise translations.}
    \label{tab:case}
\end{table*}

\subsection{Case Study}
In Table~\ref{tab:case}, we conduct a case study of \texttt{GPT-4o} and \texttt{Qwen-72B-Instruct} to further demonstrate the effectiveness of \textsc{CRAT} in improving translation accuracy, particularly in handling ambiguous or context-sensitive terms. Below is a detailed analysis of the results from these cases. 

\paragraph{(1) Handling of Polysemes} The first case illustrates the ambiguity of a term ``\emph{phin}'' in a passage related to a coffeehouse in Vietnam. ``\emph{Phin}'' refers to a traditional Vietnamese coffee brewing tool. As a polysemous term, "phin" can create confusion since its meaning is context-dependent.

\begin{itemize}
\begin{CJK}{UTF8}{gbsn}
\item  GPT-4o translates ``\emph{phin}'' as ``滤杯'' (filter bowl), which is a literally correct but contextually inappropriate translation. This misinterpretation highlights GPT-4o's inability to accurately capture the cultural significance of the term, instead of applying a literal but incorrect meaning.
\item In contrast, GPT-4o + CRAT corrects this by translating ``\emph{phin}'' as ``滴滤咖啡壶'' (drip coffee pot). 
This translation not only accurately conveys the coffee brewing method but also preserves cultural significance, highlighting CRAT's enhanced ability to manage polysemous terms with cultural relevance.
\end{CJK}
\end{itemize}

\paragraph{(2) Handling of Proper Nouns} 
``\emph{Rach Ben Nghe}'' in the first case refers to a specific canal in Ho Chi Minh City. It can be seen that:

\begin{itemize}
\begin{CJK}{UTF8}{gbsn}
\item GPT-4o retains the original name ``\emph{Rach Ben Nghe}'' in its translation, which is acceptable in some cases but can leave readers unfamiliar with Vietnamese geography without adequate context. There is no localization or explanation, making the term harder to understand for a Chinese-speaking user.
\item GPT-4o + CRAT localizes the term to "滨义河" by conducting external knowledge retrieval. This enhances the reader's comprehension by providing a more accessible translation for Chinese users.
\end{CJK}
\end{itemize}

 \paragraph{(3) Handling of Context-Dependent Terms} 
The second case addresses a more technical context, where terms like ``\emph{cosmetic overhauls}'' are discussed in the context of software updates by Apple and Google. These terms are highly domain-specific and require an understanding of both technical jargon and consumer technology language to translate accurately.
\begin{itemize}
\begin{CJK}{UTF8}{gbsn}
\item In this case, Qwen-72B-Instruct without CRAT translates ``\emph{cosmetic overhauls}'' as ``\emph{美容翻新}'' (beauty renovation), which is a misinterpretation of the technical meaning. The translation focuses on the literal meaning of ``\emph{cosmetic}'' rather than the metaphorical use of the term in a software context (i.e., user interface improvements).
\item With \texttt{CRAT},  Qwen-72B-Instruct improves its understanding by translating ``\emph{cosmetic overhauls}'' as ``\emph{外观调整}'' (appearance adjustments), which better captures the metaphorical use of the term in the software context. 
\end{CJK}
\end{itemize}

\paragraph{(4) Handling of New Terms} The last case focuses on the translation surrounding Typhoon Gaemi, highlighting the challenges LLMs face in maintaining consistent translations, as well as the importance of accurate handling of new terms.
\begin{itemize}
\begin{CJK}{UTF8}{gbsn}
\item Qwen-72B-Instruct first translates the new term ``\emph{Gaemi}'' incorrectly as ``\emph{卡米}'', and later translates it as ``\emph{盖米}'', showing inconsistency in handling the same term. Additionally, the model inaccurately renders ``\emph{Western Luzon}'' as ``\emph{西吕宋}'', which is an improper geographical reference.
\item With \texttt{CRAT},  Qwen-72B-Instruct with CRAT effectively handles this by correctly translating ``\emph{Gaemi}'' as ``\emph{格美}'' through retrieving and validating external knowledge about the emerging term.  This verifies the capabilities of CRAT to deal with emerging terms and maintain their consistency accurately, highlighting its robustness in handling real-world translation scenarios.
\end{CJK}
\end{itemize}
\section{Conclusion}
In this paper, we introduce \textsc{CRAT}, a novel multi-agent framework designed to enhance translation quality by addressing the challenges posed by context-sensitive and emerging terms. Traditional LLM-based translation models often struggle with these terms, leading to inconsistencies, especially when dealing with domain-specific vocabulary or newly coined words. Our proposed approach leverages Retrieval-Augmented Generation (RAG) combined with a causality-enhanced self-reflection mechanism, enabling the model to autonomously identify and clarify these unknown terms with greater accuracy.
The framework consists of several specialized agents, each playing a crucial role in ensuring precise translations. By detecting unknown terms, constructing knowledge graphs, and verifying the term meanings through causality-enhanced reflection, \textsc{CRAT} mitigates common translation pitfalls and reduces reliance on manual interventions. Our experimental results demonstrate that \textsc{CRAT} significantly improves translation accuracy, particularly for context-dependent terms, while maintaining consistency across different translation scenarios.

\section{Limitations}
While \textsc{CRAT} demonstrates notable improvements in translation accuracy and consistency, it has several limitations that warrant further exploration. First, the reliance on external knowledge sources during the retrieval process can introduce challenges, particularly when external data is incomplete, outdated, or contains conflicting information, which may lead to translation errors or hallucinations. Additionally, the performance of the framework is dependent on the quality and comprehensiveness of the knowledge graphs constructed, which might not always capture the full context or nuances of certain specialized terms.

Another limitation is the computational complexity introduced by the multi-agent framework, particularly in real-time translation scenarios where speed and efficiency are critical. The need for multiple agents to process, retrieve, and validate information can slow down translation times compared to more traditional, end-to-end models. Lastly, while the framework improves accuracy for unknown and context-sensitive terms, its efficacy in handling highly ambiguous or culturally specific terms still requires further investigation, especially in languages with significant variations in dialects or regional usage. Addressing these challenges will be crucial for the broader application of \textsc{CRAT} in practical, large-scale translation tasks.
\bibliography{custom}
\clearpage

\end{document}